\documentclass[10pt,twocolumn,letterpaper]{article}

\usepackage{wacv}

\usepackage{graphicx}
\usepackage{float}

\makeatletter
\@namedef{ver@everyshi.sty}{}
\makeatother
\usepackage{tikz}
\usepackage{comment}
\usepackage{amsmath,amssymb} 
\usepackage[accsupp]{axessibility}  

\usepackage{pifont}
\usepackage[acronym]{glossaries}
\usepackage{booktabs}
\usepackage{multirow}
\usepackage{multicol}

\usepackage{overpic}
\usepackage{enumitem} 
\usepackage{overpic} 
\usepackage{color}

\definecolor{turquoise}{cmyk}{0.65,0,0.1,0.3}
\definecolor{purple}{rgb}{0.65,0,0.65}
\definecolor{dark_green}{rgb}{0, 0.5, 0}
\definecolor{orange}{rgb}{0.8, 0.6, 0.2}
\definecolor{red}{rgb}{0.8, 0.2, 0.2}
\definecolor{darkred}{rgb}{0.6, 0.1, 0.05}
\definecolor{blueish}{rgb}{0.0, 0.3, .6}
\definecolor{light_gray}{rgb}{0.7, 0.7, .7}
\definecolor{pink}{rgb}{1, 0, 1}
\definecolor{greyblue}{rgb}{0.25, 0.25, 1}

\usepackage{blindtext}

\renewcommand{\paragraph}[1]{\vspace{1em}\noindent\textbf{#1}.}

\usepackage{wacv}
\usepackage{times}
\usepackage{epsfig}

\ifwacvfinal
\usepackage[breaklinks=true,bookmarks=false]{hyperref}
\else
\usepackage[pagebackref=true,breaklinks=true,colorlinks,bookmarks=false]{hyperref}
\fi

\usepackage[capitalize]{cleveref}
\crefname{section}{Sec.}{Secs.}
\Crefname{section}{Section}{Sections}
\Crefname{table}{Table}{Tables}
\crefname{table}{Tab.}{Tabs.}

\begin{document}
\title{UPAR: Unified Pedestrian Attribute Recognition and Person Retrieval}

\author{Andreas Specker\textsuperscript{1,2,3} \quad Mickael Cormier\textsuperscript{1,2,3} \quad  
Jürgen Beyerer\textsuperscript{2,1,3}\\
\textsuperscript{1}Karlsruhe Institute of Technology \quad \textsuperscript{2}Fraunhofer IOSB \quad \textsuperscript{3}Fraunhofer Center for Machine Learning \\
{\tt\small \{firstname.lastname\}@iosb.fraunhofer.de}}

%
 
\def\wacvPaperID{xxxx} 


\wacvfinalcopy 


\maketitle
\newacronym{upar}{UPAR}{Unified Pedestrian Attribute Recognition and Person Retrieval}
\newacronym{par}{PAR}{Person Attribute Recognition}
\newacronym{nlp}{NLP}{Natural Language Processing}
\newacronym{ma}{mA}{Mean Accuracy}
\newacronym{pvtv2b2}{PVTv2}{PVTv2}
\newacronym{pvtv2b2}{PVTv2-B2}{PVTv2-B2}
\newacronym{resnet50}{ResNet50}{ResNet50}
\newacronym{cn}{ConvNeXt}{ConvNeXt}
\newacronym{cnb}{ConvNeXt-B}{ConvNeXt-B}
\newacronym{swin}{Swin}{Swin}
\newacronym{swins}{Swin-S}{Swin-S}
\newacronym{peta}{PETA}{PETA}
\newacronym{pa100k}{PA100K}{PA100K}
\newacronym{rap}{RAPv2}{RAPv2}
\newacronym{market}{Market1501}{Market1501}
\newacronym{ma}{mA}{Mean Accuracy}
\newacronym{map}{mAP}{Mean Average Precision}
\newacronym{f1}{F1}{F1 score}
\newacronym{ma}{mA}{Mean Accuracy}
\newacronym{r1}{R-1}{Rank-1 Accuracy}
\newacronym{loo}{LOO}{Leave-one-out}
\newacronym{cv}{CV}{Cross-validation}
\newacronym{ema}{EMA}{Exponential Moving Average}
\newacronym{lda}{LDA}{Linear Discriminant Analysis}
\newacronym{re}{RE}{Random Erasing}
\newacronym{am}{AM}{AugMix}

\begin{abstract}
Recognizing soft-biometric pedestrian attributes is essential in video surveillance and fashion retrieval.
Recent works show promising results on single datasets. 
Nevertheless, the generalization ability of these methods under different attribute distributions, viewpoints, varying illumination, and low resolutions remains rarely understood due to strong biases and varying attributes in current datasets.
To close this gap and support a systematic investigation, we present UPAR, the Unified Person Attribute Recognition Dataset.
It is based on four well-known person attribute recognition datasets: \acrshort{pa100k}, \acrshort{peta}, \acrshort{rap}, and \acrshort{market}. 
We unify those datasets by providing 3,3M additional annotations to harmonize 40 important binary attributes over 12 attribute categories across the datasets. 
We thus enable research on generalizable pedestrian attribute recognition as well as attribute-based person retrieval for the first time.
Due to the vast variance of the image distribution, pedestrian pose, scale, and occlusion, existing approaches are greatly challenged both in terms of accuracy and efficiency. 
Furthermore, we develop a strong baseline for PAR and attribute-based person retrieval based on a thorough analysis of regularization methods.
Our models achieve state-of-the-art performance in cross-domain and specialization settings on PA100k, PETA, RAPv2, Market1501-Attributes, and UPAR.
We believe UPAR and our strong baseline will contribute to the artificial intelligence community and promote research on large-scale, generalizable attribute recognition systems. 
The code and dataset will be made available upon publication on \url{github.com}
\end{abstract}
\section{Introduction}
\label{sec:intro}

\begin{figure}[t]
\begin{center}
\includegraphics[width=0.8\linewidth]{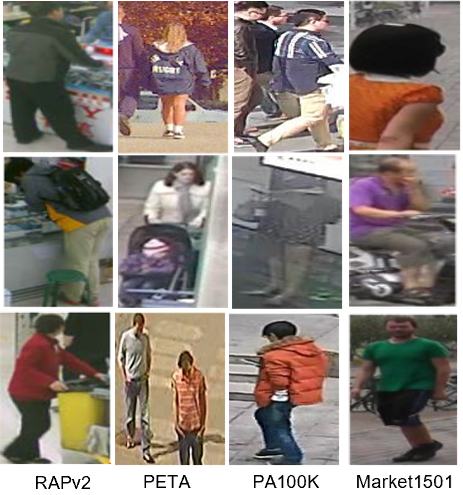}
\vspace{-0.3cm}
\end{center}
\caption{
\textbf{\acrshort{par} datasets -- }
Sample images from different \acrshort{par} datasets~\cite{deng2014pedestrian,liu2017hydraplus,lin2019improving}. 
Each dataset shows different characteristics and scenarios. 
However, meaningful out-of-distribution evaluation was impossible due to a low number of common attributes across the datasets.
Our \acrshort{upar} dataset unifies 40 attributes and thus enables cross-domain investigations.
}

\label{fig:teaser}
\vspace{-0.3cm}
\end{figure}
Pedestrian analysis poses an essential task due to the fastly growing demand for intelligent video surveillance as well as online retailing.
Relevant sub-tasks are PAR, which aims to determine the semantic attributes of people given a person image, and attribute-based person retrieval, which searches for images that show persons with the specified query characteristics.
Semantic descriptions of a pedestrian's attributes may serve as queries for attribute-based person retrieval or complementary information in common image-based re-identification in visual surveillance scenarios. 
In online retailing, attributes enable precisely filtering search parameters for clothing without requiring query images. 

In recent years,  deep learning-based visual models have shown impressive results in various tasks, and also in \acrshort{par}~\cite{chen2021enhance, tan2020relation, jia2021spatial}. 
These models heavily rely on large-scale and well-annotated datasets. 
\acrshort{par} datasets~\cite{wang2022pedestrian} offer different attribute annotations with different granularity. 
However, only a few attributes are identical across different datasets, which results in a lack of generalization analysis of \acrshort{par} at a large scale.
Moreover, the datasets are strongly biased regarding the captured scenarios.
Some datasets only include indoor or outdoor scenes, the same weather conditions, a limited number of different cameras, and only one ethnicity.
As a result, models trained on these datasets tend to overfit and may not generalize well to out-of-distribution domains.

These limitations clearly indicate the demand for a large-scale and more comprehensive benchmark for generalizable \acrshort{par}.
To this aim, we construct the \acrfull{upar} dataset which is composed of the four popular datasets \acrshort{pa100k}~\cite{liu2017hydraplus}, \acrshort{peta}~\cite{deng2014pedestrian}, \acrshort{rap}~\cite{8510891}, and \acrshort{market}~\cite{lin2019improving}.
We unify those datasets by providing additional annotations to harmonize 40 important binary attributes over 12 categories and thus bring a large number of images covering extensive diversities w.r.t human subjects, data modalities, and capturing environments, as shown in~\cref{fig:teaser}.

A total of 3,3M new binary color annotations are contributed to this dataset.

Besides extending the single datasets with previously unavailable attributes, we also enable detailed generalization evaluation for the first time. 
We propose two different evaluation protocols that represent different generalization scenarios.
The first one includes multiple training datasets, while the more difficult one only allows one dataset for training and requires evaluation on several different target domains.

In addition to introducing the \acrshort{upar} dataset, we evaluate the performance of current CNN-based models on the unified dataset.
However, due to the number of attributes and the diversity of the dataset, generalizable pedestrian attribute recognition retrieval is challenging. 
Hence, we thoroughly investigate design choices to construct a new strong baseline that achieves not only competitive results on the typical \acrshort{par} task but also concerning attribute-based person retrieval on the \acrshort{upar} dataset.

\paragraph{Contributions}
\acrshort{upar} contributes a standardized dataset to the community to launch investigations on realistic \acrshort{par} in large-scale real-world \acrshort{par} scenarios. 
Our contributions are summarized as follows.
\begin{itemize}[leftmargin=*]
\setlength\itemsep{-.3em}
\item We propose a new unified dataset with 3,3M additional annotations to harmonize 40 important attributes across four datasets. 
\acrshort{upar} is the first dataset that provides sufficient images and annotations to perform generalization experiments for \acrshort{par} and attribute-based retrieval.
\item We conduct a thorough analysis regarding design choices for generalizable PAR and attribute-based retrieval and therefore provide detailed insights.
\item We report state-of-the-art results on five \acrshort{par} datasets and outperform more complex state-of-the-art approaches concerning PAR and attribute-based retrieval.
\end{itemize}

\section{Related Work}
\label{sec:related}
\paragraph{Pedestrian Attribute Recognition}
Research on \acrshort{par} has shown significant progress over the last years.
Early deep learning-based works~\cite{li2015deepmar} propose to consider the task a multi-label classification task.
Moreover, the authors tackle the problem of imbalanced attribute distributions in the training dataset by introducing a weighted cross-entropy loss, which increases the loss for rare attributes.
The JRL~\cite{wang2017attribute} method pursues a different approach and aims at leveraging attribute correlations through long-short-term-memory modules.
A widely studied research direction is the use of attention mechanisms to focus on relevant regions and improve recognition of fine-grained attributes~\cite{liu2017hydraplus, liu2018localization, li2018pose, yang2019towards, TCLNet, sarafianos2018deep}.

It is also common to leverage features at multiple scales~\cite{yu2016weakly, sarafianos2018deep, tang2019improving, yang2020hierarchical, zhong2021improving}.

Most of the works build on large networks, which deteriorates practical applicability.
Therefore, \cite{jia2021rethinking, jia2021spatial} build on simple baseline models to achieve state-of-the-art \acrshort{par} performance.
While \cite{jia2021rethinking} introduces a strong baseline,~\cite{jia2021spatial} applies spatial and semantic regularization by exploiting that the same attribute is localized on similar positions across different images and attributes from the same semantic class with different sub-classes share similar visual features. 

These works provide impressive results for yet single datasets~\cite{lin2019improving,liu2017hydraplus,deng2014pedestrian,li2015deepmar,8510891}, with high focus on one type of scenario, \eg outdoor~\cite{lin2019improving,liu2017hydraplus} or indoor~\cite{li2015deepmar,8510891}, on comparatively medium-sized datasets~\cite{lin2019improving,deng2014pedestrian,li2015deepmar}. However, only a few attributes have an equivalent definition across different datasets such as gender, upper-body clothing length or backpack and thus enable experiments with respect to generalization on unseen domains.
Therefore researchers only have few options to document the generalization ability of their proposed approaches.
In this work, we provide the first unified dataset to perform generalization experiments at a large scale for 40 binary attributes and diverse scenarios.

\paragraph{Person Retrieval}
Researchers approach the task of attribute-based person retrieval in different ways, either by \acrshort{par} or by learning a cross-modal feature space.
Methods that follow the first approach train a \acrshort{par} model and compare the predicted attribute confidence scores with the query attribute description~\cite{vaquero2009attribute, scheirer2012multi, li2015multi, schumann2018attribute, specker2020evaluation, specker2021improving}.
This approach provides semantics, and thus retrieval results are explainable. 
However, recognizing fine-grained, local attributes in low-resolution surveillance imagery is challenging.
Further methods align attribute descriptions and image embeddings in a shared cross-modal feature space.
Approaches from the literature solve this by using high-dimensional hierarchical embeddings and an additional matching network~\cite{dong2019person}.
Further works aim to match person attributes and images in a joint feature space~\cite{yin2017adversarial, cao2020symbiotic}.
Adversarial training is applied to align the different modalities.
Jeong et al.~\cite{jeong2021asmr} argue that this procedure is often unstable and challenging due to the min-max optimization procedure.
Thus, they propose an approach that does not employ adversarial training but introduces a modality alignment loss function and a semantic regularization loss to leverage the relations between different attribute combinations explicitly~\cite{jeong2021asmr}.

Results in previous works~\cite{jeong2021asmr} indicate that learning a shared embedding space between attributes and images achieves better performance for the attribute-based retrieval task.
In contrast, our work clearly shows that state-of-the-art performance can be achieved by simply using a strong attribute classifier. 
Furthermore, semantics and explainability are provided and are often required from a real-world and application point of view.

\section{The UPAR Dataset}

Publicly available \acrshort{par} datasets are heavily biased but also limited with respect to the diversity of the captured scenes.
Except for \acrshort{pa100k}~\cite{liu2017hydraplus} and \acrshort{rap}~\cite{8510891}, most of the existing public \acrshort{par} datasets include less than 50,000 annotated pedestrians in either outdoor or indoor settings.
For example, \acrshort{market} only contains outdoor imagery captured during summer on campus.
As a result, attributes are biased towards young Asian people wearing short summer clothing and thus do not reflect reality well.
For this reason, findings on such datasets are limited and can hardly be transferred to other scenarios or applied in real-world applications.
There are significant distribution shifts concerning attribute distributions as well as images across the datasets.
\begin{figure}
\begin{center}
\includegraphics[width=.9\linewidth,trim={0 1.2cm 0 1.2cm},clip]{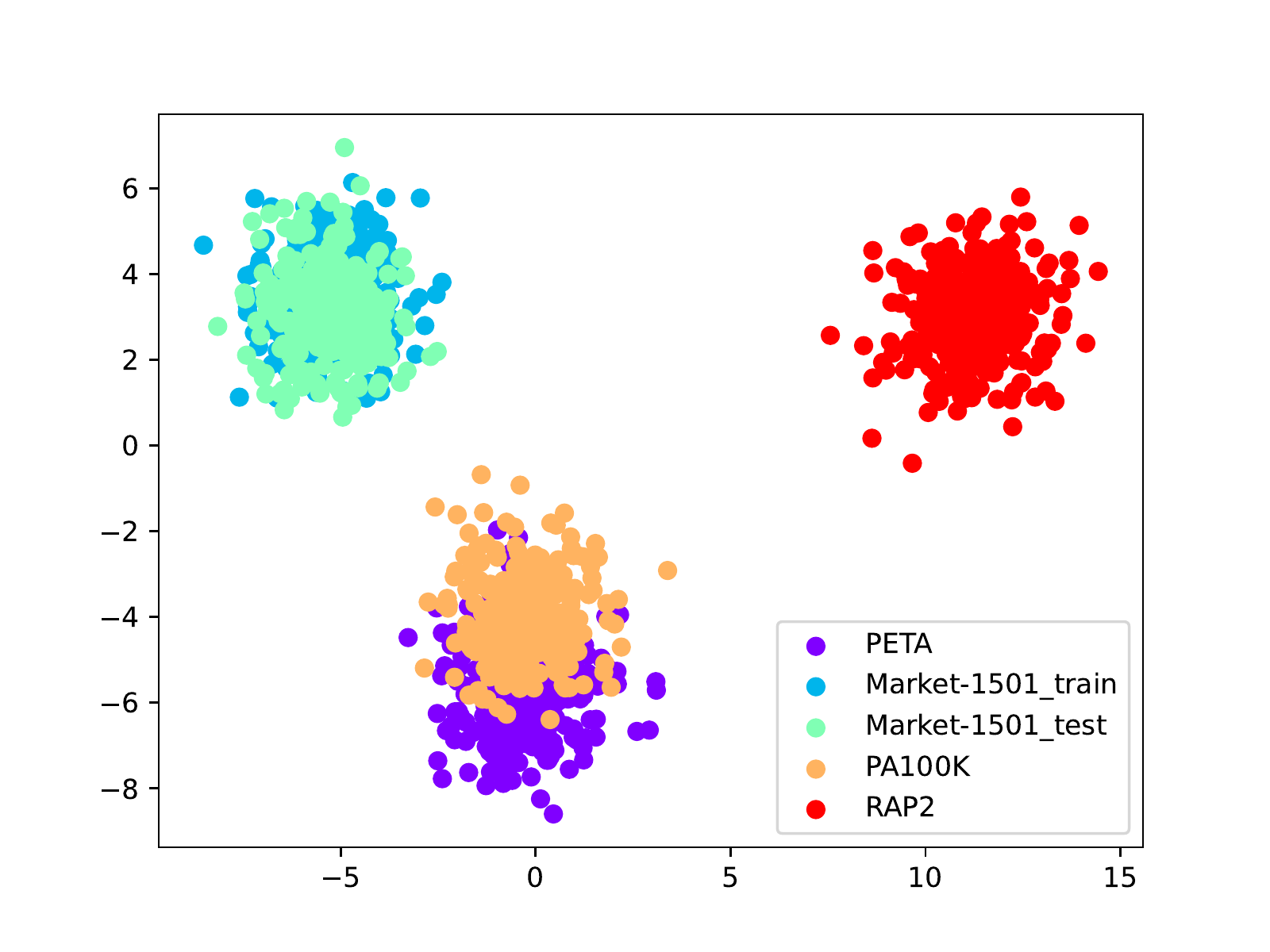}
\end{center}
\caption{
\textbf{UPAR Domains} -- Distribution of image embeddings extracted using an ImageNet pre-trained Inception model.
Train and test set embeddings from the same dataset overlap entirely. In contrast, our \acrshort{upar} dataset as a combination of multiple sub-datasets with disjunct data distributions poses a more realistic and challenging problem and requires models to generalize well across different domains.
}
\label{fig:domains}
\vspace{-0.2cm}
\end{figure}
The significant distribution shifts of the images are visualized in \cref{fig:domains}.
It shows the distribution of image embeddings extracted using an ImageNet pre-trained Inception model.
The outputs of the last feature layer are transformed into two-dimensional features using the \acrlong{lda} method.
One can observe that train and test set embeddings originating from the same dataset overlap entirely, thus simplifying the \acrshort{par} task greatly since models are only required to generalize well regarding comparatively small shifts of attribute distributions.
In contrast, real-world applications require generalization to out-of-distribution domains, i.e., other datasets with different characteristics in our case.
Out-of-distribution evaluation poses a much more complex and realistic task.
To enable research on such realistic settings, we propose the \acrshort{upar} dataset by unifying multiple existing \acrshort{par} datasets concerning their attribute annotations and introducing two generalization evaluation protocols.

\subsection{UPAR Composition}
We construct our \acrshort{upar} dataset by combining the images included in the four publicy available datasets \acrshort{pa100k}~\cite{liu2017hydraplus}, \acrshort{peta}~\cite{deng2014pedestrian}, \acrshort{rap}~\cite{8510891}, and \acrshort{market}~\cite{lin2019improving,zheng2015scalable}.
Due to several limitations regarding the creation of real-world surveillance datasets, particularly regarding privacy issues, we argue that already existing public datasets have much potential to be leveraged.
Those datasets provide a large diversity concerning scenarios, ethnicities, and attributes, which makes \acrshort{upar} less biased.
The \acrshort{market}-Attribute~\cite{lin2019improving,zheng2015scalable} dataset is an extended version of the \acrshort{market} dataset augmented with the annotation of 27 attributes. 
The \acrshort{market} dataset was collected outdoors in front of a supermarket using five high-resolution cameras and a low-resolution one. 
The dataset has 32,668 bounding boxes for 1,501 identities. 
\acrshort{pa100k}~\cite{liu2017hydraplus} is currently the largest \acrshort{par} dataset with 100,000 pedestrian images from various outdoor surveillance cameras with a large variance in image resolution, lighting conditions, and environments. 
The dataset provides annotation for 26 attributes.
The Richly Annotated Pedestrian (RAP)\cite{li2015deepmar,8510891} dataset is currently the largest pedestrian attribute dataset composed of indoor scenes. 
There are two versions of this dataset. 
The first version, RAPv1~\cite{li2015deepmar}, was collected from a surveillance camera in shopping malls over three months and consists of 41,585 pedestrian images and 72 annotated attributes.
\acrshort{rap}~\cite{8510891} is an extension intended as a unified benchmark for both person retrieval and \acrshort{par} in indoor real-world surveillance scenarios. 
The dataset contains 84,928 images for 2,589 person identities and 25 different scenes. 
The attributes annotated are the same as in RAPv1.
Finally, the PEdesTrian Attribute (PETA)~\cite{deng2014pedestrian} dataset combines 19,000 pedestrian images from 10 publicly available datasets with indoor and outdoor scenes and 61 binary and four multi-class attributes. 
While widely used by the community for benchmarking \acrshort{par} algorithms, those datasets have only a few attribute annotations in common,
which significantly impairs generalization- and cross-evaluation.

To enable research on generalization, we unify 40 binary attributes over 12 categories between these datasets.

\subsection{UPAR Attributes \& Statistics}
The 40 unified attributes were selected regarding their importance for person retrieval in video surveillance scenarios.
Our annotations cover a variety of different attribute types to allow broad investigations.
Global-scale attributes such as age or gender and small-scale attributes, e.g., glasses, are included in the dataset.
There are attributes belonging to the twelve different categories age, gender, hair length, upper-body clothing length, upper-body clothing color, lower-body clothing length, lower-body clothing color, lower-body clothing type, accessory-backpack, accessory-bag, accessory-glasses and accessory hat. 
A complete list of the attributes can be found in the supplementary material. 

As a result, after cleaning images from the original datasets that could not be satisfyingly annotated, the \acrshort{upar} dataset contains 224,737 images with annotations for 40 binary attributes over 12 categories. 
The dataset follows the splits from the original datasets for a total of 148,048 train, 30,830 validation, and 45,859 test images.
We provide annotations for 11 unique colors (plus other and mixture) for 100,000 images and two attributes for a total of 2,57M binary color annotations.
Furthermore, we provide annotation for the lower-body clothing length for the \acrshort{peta} and \acrshort{rap} datasets. Annotation for glasses are contributed to \acrshort{pa100k}, \acrshort{peta} and \acrshort{market}, age attribute to \acrshort{peta} and hair length to \acrshort{pa100k} and \acrshort{market} resulting in further 0.8M new annotations, for a total of 3,3M manually labled and validated new binary annotatinos.
With these new annotations, we enable multiple splits for leave-one-out evaluation to assess generalizability (see~\cref{tab:upar_splits}).

Moreover, the \acrshort{upar} can be used as a whole for the evaluation of specialization approaches.

With our dataset, we attempt to mitigate several issues regarding \acrshort{par} datasets.

First, these are highly unbalanced regarding some attributes. 

In \acrshort{upar} even less represented attributes show at least 446 examples. It is a step in the right direction, however, the gap between the less and best represented attributes remains significant.
The reason for this is the data context and camera configuration in which the datasets were captured. 
The environment (market, mall, campus, etc.) greatly impacts the distribution of clothing and demographics. Publicly available datasets tend to have biases towards ethnicity and cultures, \eg mostly representing western or south-east Asian cultures. 
We slightly mitigate this by unifying four large datasets. 
Future works should focus explicitly on the representation of larger and more diverse communities with diverse clothing.
While not specifically addressing this, the increased diversity of our dataset tends to mitigate the involuntary learning of biometrics. 
Second, researchers are confronted with privacy issues and human rights violations when gathering a real-world dataset. 
From a research point of view, \acrshort{par} datasets should be captured by real-world surveillance cameras, with subjects unaware of being filmed and with behavior as natural as possible. 
From an ethical perspective, however, pedestrians should consent before their images are annotated and publicly distributed. 
This important dilemma also touches already existing datasets. 
For example, the Duke-MTMC~\cite{Ristani2016PerformanceMA} and Celeb1M~\cite{Guo2016MSCeleb1MAD} datasets were withdrawn due to privacy violation issues.
In this paper, we argue that already existing datasets are enough for our purposes and that these are not used to their full potential due to lacking annotation. 
Therefore, we believe that extending already existing and widely accepted datasets instead of acquiring more personal data is a viable strategy for scenarios such as \acrshort{par}.


\subsection{Data Annotation Process}
The \acrshort{pa100k} dataset does not provide color annotation, which is the reason why we asked $16$ paid annotators to manually define the color of the upper-body clothing and lower-body clothing for eleven unique colors plus additional classes to indicate multiple colors or colors not included in the color list. 
The annotation is performed for each image individually, \ie, when describing the attributes of the same person in different images, only the visible information is taken into account.
\begin{figure}[t]
\begin{center}
\includegraphics[width=0.7\linewidth]{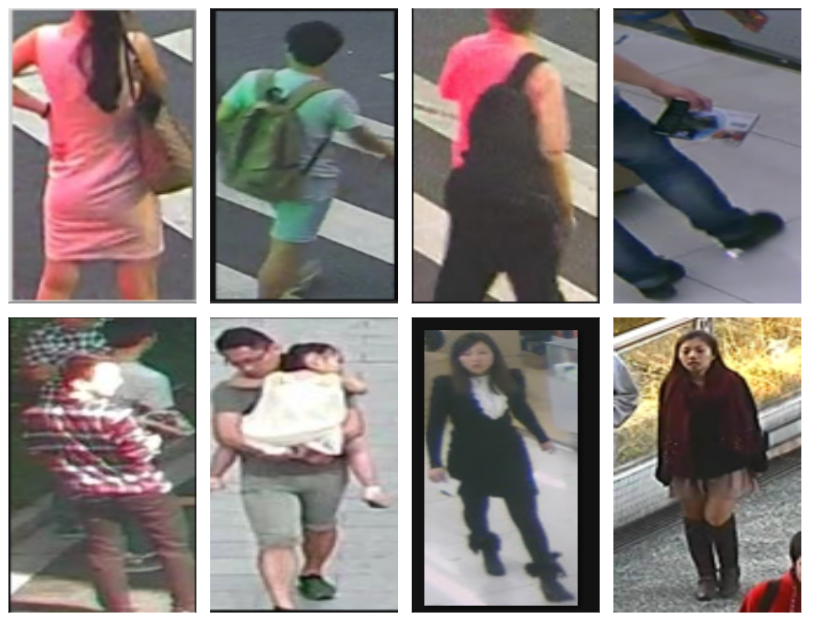}
\vspace{-0.3cm}
\end{center}
\caption{
\textbf{Annotation challenges -- } For the first three images, the real color of the clothing differs from the color seen. In the fourth image, the upper body is missing. Sometimes the primary person is difficult to tell, as can be seen in the first two images of the second row. Do leggings make the lower-body clothing long? (yes) Do long boots make the lower body clothing long? (no)
}
\label{fig:anno_challenges}
\vspace{-0.3cm}
\end{figure}
Thus, different attribute values may be assigned to the same person in different images depending on external factors such as lighting, shadows cast by objects or persons, or the clothing area visible in the frame, as illustrated in~\cref{fig:anno_challenges}.
In general, annotators are confronted with a variety of challenges. 
For instance, some images depict multiple persons and, thus, it is often difficult to define the primary person in the image due to occlusion. 
Additionally, colors might not be unambiguously assignable to unique color classes. 
In such cases, images are assigned to the "other" collection class, which, among others, includes metallic colors or the color beige. 
If the corresponding body area of the primary person is not visible in the frame or if the primary person is depicted in false colors, e.g. grayscale, an assignment to the class unknown is made, and the image is discarded from the \acrshort{upar} dataset. For more info, we refer to the supplementary materials.


\subsection{Benchmark and Metrics}
We exploit two schemes to assess generalization ability on UPAR: \acrfull{cv} and \acrfull{loo}.
\begin{table}[t]
\centering
\resizebox{\linewidth}{!}{ 
\begin{tabular}{@{}l|cc|cc@{}}
\toprule
\multirow{2}{*}{Split ID} & \multicolumn{2}{c|}{\acrlong{loo}} & \multicolumn{2}{c}{\acrlong{cv}} \\ 
& Training & Evaluation & Training & Evaluation \\
\midrule
0 & \acrshort{pa100k}, PETA, \acrshort{rap} & \acrshort{market} & \acrshort{market} & \acrshort{pa100k}, PETA, \acrshort{rap} \\
1 & \acrshort{market}, PETA, \acrshort{rap} & \acrshort{pa100k} & \acrshort{pa100k} & \acrshort{market}, PETA, \acrshort{rap} \\
2 & \acrshort{market}, \acrshort{pa100k}, \acrshort{rap} & PETA & PETA & \acrshort{market}, \acrshort{pa100k}, \acrshort{rap} \\
3 & \acrshort{market}, \acrshort{pa100k}, PETA & \acrshort{rap} & \acrshort{rap} & \acrshort{market}, \acrshort{pa100k}, PETA \\
\bottomrule
\end{tabular}
} 
\caption{
\textbf{\acrshort{upar} splits} -- Split definitions for the two \acrshort{upar} generalization evaluation schemes.
\acrfull{cv} protocol is more challenging since only sub-dataset is used for training.
In both cases, evaluation is performed on unseen domains.
} 
\label{tab:upar_splits}
\end{table}
Different training and testing splits are given in~\cref{tab:upar_splits}.

The \acrshort{cv} protocol is the more challenging one since it measures the generalization performance on multiple domains given training data from a single distribution.
Metrics are calculated by evaluating the test datasets separately, computing the average across these test datasets for each split, and finally determining the mean and standard deviation across the splits.  
This procedure ensures that the datasets contribute equally to the final results and not proportionally to their number of test images.

The second generalization evaluation scheme uses \acrshort{loo} to simulate the case when training data from several domains are available, but data from the target distribution is missing.
I.e., three of the four sub-datasets are leveraged for training and the remaining one for evaluation.
Similar to the first protocol, there are four splits, and each sub-dataset is used for testing once.
The final evaluation scores are the average and standard deviation across the four splits.

Last but not least, the dataset can be used in total to perform evaluation analogous to the original datasets.
In this case, the training images and the test images from all sub-datasets are combined.

Typically, label-based and instance-based metrics are used to evaluate \acrshort{par} approaches~\cite{8510891}.
Label-based metrics, such as the \acrlong{ma}, focus on the recognition accuracy of single attributes.
The classification accuracy is computed separately for positive and negative samples of the attributes, and the mean is calculated for each attribute.
Then, the \acrshort{ma} is the average across all attributes.
In contrast, instance-based metrics, i.e., the F1 score in our case, focus on the attribute description assigned to persons.
The precision and recall values for each test set sample is calculated as the basis for the F1 score, which is the harmonic mean of those values.
In this case, the average is computed over all instances in the test set.
Concerning attribute-based person retrieval, we rely on the typically used \acrfull{map} and \acrfull{r1} metrics.
While the \acrshort{map} reflects the quality of the entire retrieval ranking, the \acrshort{r1} score is the accuracy of the first position in the ranking.

Note that attributes without a positive sample in the training data of the split are excluded.
Furthermore, attribute queries for the retrieval are determined by computing all unique sets of attributes occurring in the respective test set.

\section{Proposed Strong Baseline}

This work aims to develop a strong baseline universally applicable to the \acrshort{par} and attribute-based person retrieval tasks.
In contrast to \acrshort{par}, there is no simple and strong baseline specifically tailored and evaluated for attribute-based person retrieval to achieve state-of-the-art comparable performance.
Hence, we conduct a detailed analysis of different design aspects to construct a baseline approach that achieves state-of-the-art performance on \acrshort{upar} as well as all sub-datasets for both tasks.
We apply current backbone models and optimizers, improve the training batch size and leverage several regularization methods resulting in an approach robust against overfitting and overconfidence.

\vspace{-0.3cm}
\paragraph{Architecture}
We follow the typical baseline architecture, which consists of a backbone model with a fully-connected classification head.
We make use of only one classification layer that follows the global pooling layer of the backbone.
Our experiments showed that average pooling works best, which is why it is applied throughout all investigations.

\vspace{-0.3cm}
\paragraph{Backbones}
Most of the approaches still rely on \acrshort{resnet50}~\cite{he2016deep} as the backbone model since it achieved competitive performance at reasonable computational costs on various tasks for years.
However, several new network architectures have been proposed recently.
Starting with the ViT~\cite{dosovitskiy2021image}, transformer-based models gained increasing importance and outperformed CNNs in various research fields. 
Furthermore, improved CNN models are introduced, e.g., the \acrshort{cn}~\cite{liu2022convnet} architecture which resembles the structure of modern vision transformers but with convolutional blocks.
In our work, we compare the \acrshort{resnet50}~\cite{he2016deep} model with vision transformers \acrshort{swins}~\cite{liu2021Swin} and \acrshort{pvtv2b2}~\cite{wang2021pvtv2}, and \acrshort{cnb}~\cite{liu2022convnet}.

\vspace{-0.3cm}
\paragraph{\acrlong{ema}}
Research on various model architectures such as EfficientNet~\cite{tan2019efficientnet} has shown that using \acrshort{ema}s of a model's trainable parameters for evaluation may significantly improve the results.
An additional \acrshort{ema} model is maintained during training. 
In each iteration, the original model is updated first.
Subsequently, the \acrshort{ema} is computed for each parameter based on the current \acrshort{ema} values and the new parameter values of the original model.

\vspace{-0.3cm}
\paragraph{Batch size}
Our experimental investigations indicate that the training batch size influences the final evaluation performance.
The reason is that if the batch size is too large, rarely occurring attributes only have little impact on batch updates of the model's parameters.
However, if the batch size is small, the model tends to overfit on some samples, which may result in unstable training behavior and poor generalization performance.
Furthermore, regularizations such as \acrshort{ema} may require fewer images per batch to learn meaningful features for fine-grained or rare attributes, especially on small datasets.

\vspace{-0.3cm}
\paragraph{Dropout}
Dropout is a well-studied regularization method to avoid overfitting.
In particular, this improves the generalization ability on data from unseen domains, i.e., the results on the \acrshort{upar} evaluation protocols.
However, it is also beneficial for specialization tasks within the same domain since attribute datasets include huge biases, and models start to overfit after a few epochs.

\vspace{-0.3cm}
\paragraph{Label smoothing}
Two common problems with deep learning-based approaches for image classification tasks are overconfidence and overfitting.
Applying label smoothing tackles both problems simultaneously.
If a model is overconfident, output predictions do not reflect the accuracy, i.e., output scores for frequently occurring attributes may be significantly higher than the accuracy.

Typically, label smoothing is applied when the cross-entropy loss is used as an optimization target for multi-class classification.
However, we adopt the same idea to the multi-label attribute recognition problem as follows:
\begin{equation}
\label{eq:ls}
    y_{ls} = (1 - \alpha) * y + \alpha * (1 - y) .
\end{equation}
In Equation~\ref{eq:ls}, $y$ is the original, binary attribute label, $\alpha$ represents the hyperparameter to control the impact of label smoothing, and $y_{ls}$ stands for the resulting smoothed training label.

\vspace{-0.3cm}
\paragraph{Optimizer}
Many works use adaptive optimizers such as Adam to train their neural networks since they require less fine-tuning of hyperparameters and typically lead to faster training than stochastic gradient descent.
However, models trained with Adam have shown to generalize worse.
Loschilov \etal~\cite{loshchilov2018fixing} indicate that the reason is the less effective L2 weight regularization for adaptive optimization methods and proposed the AdamW optimizer, which fixed the problem.
Nevertheless, current \acrshort{par} baselines~\cite{jia2021rethinking} still rely on the vanilla version of Adam and therefore suffer from poor generalization.
In contrast, we apply the AdamW optimizer in our approach and show significant improvement.
Other algorithms, e.g., RAdam~\cite{liu2019variance}, were investigated but achieved worse results than AdamW.

\vspace{-0.3cm}
\paragraph{Data augmentation}
Another commonly used method to reduce overfitting is data augmentation.
We conducted experiments with various augmentation methods and finally found that \acrfull{re}~\cite{zhong2020random} and \acrfull{am}~\cite{hendrycks2019augmix} augmentation algorithms improve the results.
Besides, random flipping and cropping established themselves as standard augmentation methods and are thus always applied in our experiments.

\section{Experiments}

In this section, we present and discuss experimental results.
First, general information about training and evaluation is given, followed by thorough studies regarding generalization on the \acrshort{upar} dataset.
Second, we compare our baseline model to state-of-the-art methods on single datasets.

\subsection{Training Setup}
In our work, we leverage the work of Jia \etal~\cite{jia2021rethinking} as our baseline.
We train all models using a learning rate of $1e-4$, weight decay of $5e-4$, and the Adam optimizer if otherwise stated.
Regarding the learning rate schedule, we apply a plateau scheduler that reduces the learning rate by a factor of 0.1 if the validation results do not improve for four epochs.
Batches of size 64 are used per default, and gradient clipping is applied after computing the gradients based on the weighted cross-entropy loss function~\cite{li2015deepmar}.
The backbone networks are initialized with pre-trained weights from the ImageNet dataset~\cite{deng2009imagenet}.
More details can be found in the supplementary material.

We optimize all models in this work, focusing on retrieval performance since it is the relevant measure for most applications.
Hence, obtained \acrshort{par} results are not optimal, especially concerning the \acrshort{ma}.
However, we are still able to outperform state-of-the-art approaches.

\subsection{UPAR Generalization}
\vspace{-0.3cm}

\paragraph{CV Results}
Results for training on one sub-dataset and evaluating on the others are provided in~\cref{tab:upar_cv}.
\begin{table}[t]
\centering
	\resizebox{\linewidth}{!}{ 
		\begin{tabular}{@{}l|l|cccc@{}}%
			\toprule%
			Method & Backbone & \acrshort{ma}                     & \acrshort{f1}   &   \acrshort{map}   &   \acrshort{r1} \\%
			\midrule%
			VAC \cite{guo2019visual}$\dagger$ & \acrshort{resnet50} & $64.3 \pm 1.8$ & $71.4 \pm 6.1$ & $6.2 \pm 3.3$ & $7.4 \pm 4.1$ \\%
            ALM \cite{tang2019improving}$\dagger$  & BN-Inception & $66.3 \pm 2.0$ & $71.0 \pm 5.4$ & $5.5 \pm 3.0$ & $7.3 \pm 4.0$ \\%
            SAL~\cite{cao2020symbiotic}$\dagger$ & \acrshort{resnet50} & -- & -- & $3.7 \pm 1.5$ & $4.8 \pm 2.0$ \\%
			\midrule%
			\multirow{4}{*}{Baseline~\cite{jia2021rethinking}$\dagger$}                   & \acrshort{resnet50}      & $65.5 \pm 2.2$ & $71.2 \pm 5.6$ & $6.6 \pm 3.4$ & $8.7 \pm 4.5$ \\%
			
				& \acrshort{swins} & $68.4 \pm 2.2$ & $74.5 \pm 5.2$ & $8.6 \pm 3.5$ & $9.9 \pm 4.5$ \\%
				& \acrshort{pvtv2b2} & $68.1 \pm 2.1$ & $75.0 \pm 4.2$ & $8.6 \pm 3.4$ & $10.4 \pm 4.5$ \\%
				& \acrshort{cnb} & $70.2 \pm 1.2$ & $76.8 \pm 4.2$ & $11.1 \pm 3.9$ & $12.9 \pm 5.1$ \\%
			\midrule%
			+ EMA & \acrshort{cnb} & $69.3 \pm 1.0$ & $78.0 \pm 3.5$ & $11.8 \pm 4.3$ & $13.1 \pm 5.1$ \\%
			+ optimal BS                                 & \acrshort{cnb}          & $69.4 \pm 1.1$ & $78.3 \pm 3.0$ & $12.0 \pm 4.3$ & $13.4 \pm 5.3$ \\%
			+ dropout & \acrshort{cnb} & $69.8 \pm 1.4$ & $79.1 \pm 3.1 $ & $12.6 \pm 4.2$ & $14.3 \pm 4.9$ \\%
			+ label smoothing & \acrshort{cnb} & $70.0 \pm 1.5$ & $79.2 \pm 3.0 $ & $12.7 \pm 4.0$ & $14.5 \pm 4.7$  \\%
			+ AdamW & \acrshort{cnb} & $\mathbf{70.8}\pm 1.7$ & $79.5 \pm 3.1 $ & $13.4 \pm 4.3$ & $15.4 \pm 5.4$ \\%
			+ AugMix (Ours) & \acrshort{cnb} & $70.5 \pm 1.9$ & $\mathbf{80.1} \pm 2.7 $ & $\mathbf{13.7} \pm 4.2$ & $\mathbf{15.7} \pm 5.0$ \\%
			\midrule%
			Ours & \acrshort{resnet50} & $67.0 \pm 2.5$ & $74.2 \pm 4.5$ & $8.3 \pm 4.1$ & $10.2 \pm 5.0$ \\%
			\bottomrule%
		\end{tabular}%
	}
	\caption{
		\textbf{UPAR CV} -- Generalization results when using data from only one domain for training.
		The proposed optimizations of the baseline approach increase the performance regarding all metrics.
		$\dagger$~Results were produced using the official implementation.
	} 
	\label{tab:upar_cv}%
\end{table}

Regarding approaches from literature, one can observe that the strong baseline~\cite{jia2021rethinking}, which also serves as our baseline, achieves the best results.
Especially, more complex methods that, e.g., apply GANs~\cite{cao2020symbiotic} achieve poor results for generalizing on out-of-distribution domains.
Furthermore, using \acrshort{resnet50} as the backbone model performs significantly worse than using more recent CNN- or transformer-based models.
Best results are obtained by the \acrshort{cnb} architecture.
This applies to both tasks \acrshort{par} and retrieval.
As a result, we rely on this backbone model for our method.
All the proposed regularization techniques improve the scores.
Adding Dropout to the last feature layer of the backbone and using AdamW instead of vanilla Adam has the most considerable impact on performance.
Interesting findings are that EMA evaluation is only beneficial when enough training data is available.
Splits 0 and 2, i.e., leveraging the small \acrshort{market} or \acrshort{peta} datasets for training, achieve better results without \acrshort{ema} computation of weights.
In contrast, regularization with \acrshort{ema} requires a lower batch size of 32 in this generalization setting (splits 1 \& 3).
Otherwise, small-scale features for highly localized attributes are hardly learned.
Similar differences between splits with more and less training data are observed for the label smoothing hyperparameter $\alpha$.
The two smaller splits need more regularization, i.e., $\alpha=0.1$, while $\alpha=0.05$ is sufficient for the other splits.

It is also noteworthy that applying AdamW before adding regularization does not improve the results.
The benefits of AdamW can not compensate for the model's tendency to overfit.
Concerning data augmentation, only \acrshort{am} improves the results.
The reason is that \acrshort{re} may lead to overfitting on the training data since if relevant image regions are erased, some attributes may not be determinable, which raises the risk of recognizing attributes for this image based on irrelevant background characteristics.
In conclusion, the experiments indicate that the improvements are universal, but the parameterization highly depends on the amount of available training data when focusing on generalization to unseen domains.
If few training images ($<$15,000 images in our case) are available, more regularization will be needed, and \acrshort{ema} is not advantageous.
In contrast, training with more data ($>$50,000 images in our case) benefits from \acrshort{ema}, but the batch size should be reduced to keep the influence of images showing rare attributes.

Our final results greatly outperform the state-of-the-art and show that current algorithms with a simple baseline architecture are well suited for generalization tasks.
Finally, we apply the same bag of tricks with identical hyperparameters to the \acrshort{resnet50} backbone, although the parameters might not be optimal.
However, results indicate that there also is a significant improvement, and the state-of-the-art is outperformed.
\begin{table}[t]
\centering
	\resizebox{\linewidth}{!}{ 
		\begin{tabular}{@{}l|l|cccc@{}}%
			\toprule%
			Method & Backbone & \acrshort{ma}                     & \acrshort{f1}   &   \acrshort{map}   &   \acrshort{r1} \\%
			\midrule%
			VAC \cite{guo2019visual}$\dagger$ & \acrshort{resnet50} & $64.3 \pm 1.8$ & $71.4 \pm 6.1$ & $6.2 \pm 3.3$ & $7.4 \pm 4.1$ \\%
            ALM \cite{tang2019improving}$\dagger$  & BN-Inception & $72.0 \pm 2.3$ & $78.0 \pm 2.7$ & $11.4 \pm 4.2$ & $13.9 \pm 9.0$  \\%
            SAL~\cite{cao2020symbiotic}$\dagger$ & \acrshort{resnet50} & -- & -- & $7.9 \pm 3.4$ & $10.5 \pm 7.2$ \\%
			\midrule%
			\multirow{4}{*}{Baseline~\cite{jia2021rethinking}}                   & \acrshort{resnet50}      & $71.5 \pm 1.9$ & $78.7 \pm 2.9$ & $13.4 \pm 5.0$ & $16.0 \pm 10.5$ \\%
				& \acrshort{swins} & $72.9 \pm 1.9$ & $80.4 \pm 2.6$ & $14.6 \pm 4.4$ & $17.1 \pm 9.3$ \\%
				& \acrshort{pvtv2b2} & $72.5 \pm 1.7$ & $80.7 \pm 2.8$ & $15.0 \pm 4.4$ & $17.0 \pm 8.7$ \\%
				& \acrshort{cnb} & $73.2 \pm 1.8$ & $81.8 \pm 2.7$ & $17.2 \pm 3.7$ & $20.1 \pm 9.4$\\%
			\midrule%
			+ EMA & \acrshort{cnb} & $72.4 \pm 1.9$ & $83.1 \pm 2.1$ & $19.3 \pm 5.3$ & $21.3 \pm 9.9$\\%
			+ dropout & \acrshort{cnb} & $73.4 \pm 2.4$ & $83.5 \pm 2.2 $ & $20.2 \pm 5.2$ & $22.1 \pm 9.1$  \\%
			+ label smoothing & \acrshort{cnb} & $74.4 \pm 2.2$ & $83.5 \pm 2.1$ & $20.2 \pm 5.1$ & $22.2 \pm 8.9$ \\%
			+ AdamW & \acrshort{cnb} & $75.3\pm 2.0$ & $84.1 \pm 2.0 $ & $20.9 \pm 5.3$ & $22.9 \pm 9.5$ \\%
			+ AugMix (Ours) & \acrshort{cnb} & $\mathbf{75.4} \pm 1.9$ & $\mathbf{84.2} \pm 2.0 $ & $\mathbf{21.4} \pm 5.4$ & $\mathbf{23.4} \pm 9.1$ \\%
			\midrule%
			Ours & \acrshort{resnet50} & $72.6 \pm 2.4$ & $81.4 \pm 2.3$ & $16.0 \pm 5.4$ & $18.5 \pm 10.3$ \\%
			
			\bottomrule%
		\end{tabular}%
	}
	\caption{
		\textbf{UPAR LOO} -- Generalization results when using data from multiple domains for training.
		The proposed optimizations of the baseline approach increase the performance regarding all metrics.
		$\dagger$~Results were produced using the official implementation.
	} 
	\label{tab:upar_loo}%
\end{table}

\begin{table*}[t]
    \centering
	\resizebox{\linewidth}{!}{ 
		\begin{tabular}{@{}l|l|cccc|cccc|cccc|cccc|cccc@{}}%
			\toprule%
			\multirow{2}{*}{Method}                       & \multirow{2}{*}{Backbone}         & \multicolumn{4}{c|}{PETA} & \multicolumn{4}{c|}{PA100K} & \multicolumn{4}{c|}{RAP2} & \multicolumn{4}{c|}{Market-1501} & \multicolumn{4}{c}{UPAR} \\%
			\cline{3-22}%
			                                              &                                      & \acrshort{ma}                    & \acrshort{f1}  &   \acrshort{map}   &   \acrshort{r1}                        & \acrshort{ma}                      & \acrshort{f1}   &   \acrshort{map}   &   \acrshort{r1}                         & \acrshort{ma}                    & \acrshort{f1}       &   \acrshort{map}   &   \acrshort{r1}                      & \acrshort{ma}                     & \acrshort{f1}          &   \acrshort{map}   &   \acrshort{r1}   & \acrshort{ma}                     & \acrshort{f1}          &   \acrshort{map}   &   \acrshort{r1}             \\%
			\midrule%
			\multirow{4}{*}{Baseline~\cite{jia2021rethinking}$^\dagger$}%
			    & \acrshort{resnet50} & 84.0 & 86.3 & 20.7 & 21.3 & 81.6 & 88.1 & 23.8 & 31.1 & 77.4 & 78.5 & 17.0 & 12.1 & 76.5 & 83.6 & 23.8 & 39.5 & 83.2 & 87.3 & 21.2 & 23.5 \\%
				& \acrshort{swins} & 86.6 & 87.7 & 24.1 & 24.1 & 83.2 & 88.5 & 25.7 & 32.9 & \textbf{80.0} & 80.3 & 21.0 & 15.0 & 78.2 & 84.7 & 26.3 & 39.5 & 81.4 & 86.9 & 19.9 & 21.4 \\%
				& \acrshort{pvtv2b2} & 84.1 & 86.3 & 20.5 & 20.3 & 82.1 & 88.7 & 25.8 & 33.5 & 78.0 & 78.8 & 17.2 & 12.0 & 77.5 & 83.8 & 25.5 & 39.5 & 83.7 & 88.6 & 26.1 & 28.5 \\%
				& \acrshort{cnb} & 86.1 & 88.1 & 24.4 & 24.4 & 82.2 & 88.5 & 26.2 & 34.5 & 79.3 & 80.0 & 20.5 & 14.4 & 80.7 & 85.7 & 31.6 & 47.7 & 83.9 & 89.2 & 26.9 & 28.7 \\%
			\midrule%
			+ EMA & \acrshort{cnb} & 85.6 & 88.4 & 25.4 & 24.7 & 83.7 & 89.7 & 29.3 & 36.3 & 78.5 & 80.9 & 22.2 & 15.9 & 77.5 & 86.2 & 29.9 & 45.3 & 83.9 & 89.2 & 27.3 & 29.3 \\%
			+ optimal BS & \acrshort{cnb} & 87.0 & 88.5 & 26.7 & 26.5 & 83.7 & 89.7 & 29.3 & 36.3 & 78.5 & 80.9 & 22.2 & 15.9 & 79.6 & 86.7 & 35.7 & 51.0 & 83.9 & 89.2 & 27.3 & 29.3 \\%
			+ dropout & \acrshort{cnb} & 87.1 & 88.4 & 26.5 & 26.8 & 84.0 & 89.7 & 30.0 & 37.6 & 79.9 & 81.0 & 22.3 & 16.3 & 79.3 & 86.6 & 36.5 & 49.8 & 84.3 & 89.4 & 27.7 & 29.6 \\%
			+ label smoothing & \acrshort{cnb} & 87.4 & 88.8 & 27.7 & 27.7 & \textbf{84.8} & 90.0 & 30.0 & 38.5 & 79.5 & \textbf{81.1} & 21.7 & 15.8 & 79.0 & 86.8 & 37.4 & 50.2 & 84.9 & 89.4 & 27.9 & 30.1 \\%
			+ AdamW & \acrshort{cnb} & 88.1 & 89.7 & \textbf{30.4} & 29.3 & \textbf{84.8} & 90.2 & 30.5 & \textbf{39.5} & 79.8 & \textbf{81.1} & 20.4 & 14.9 & 81.4 & \textbf{87.7} & 40.3 & 52.5 & 85.7 & \textbf{90.2} & 30.7 & 32.7 \\%
			+ Random Erasing & \acrshort{cnb} & \textbf{88.4} & \textbf{89.9} & 30.2 & \textbf{29.7} & 84.6 & \textbf{90.4} & \textbf{30.7} & 38.8 & 79.7 & \textbf{81.1} & \textbf{22.5} & \textbf{16.4} & \textbf{81.5} & 87.6 & \textbf{40.6} & \textbf{55.4} & \textbf{85.9} & \textbf{90.2} & \textbf{31.6} & \textbf{33.9} \\%
			\bottomrule%
		\end{tabular}%
	}
	\caption{
		\textbf{Specialization results} -- Analogous to the generalization protocols on \acrshort{upar}, the proposed optimizations of the baseline method also lead to significant improvements for specializing on a single domain.
		$\dagger$ Results were produced using the official implementation.
	} 
	\label{tab:public}%
\end{table*}

\paragraph{LOO Results}
Results for the \acrshort{loo} generalization evaluation protocol are given in \cref{tab:upar_loo}.
Observations are similar to those seen in the previous section.
Since more training data is available in each split, \acrshort{ema} always led to an improvement and adjusting the batch size was unnecessary.
Again, our approach significantly outperforms methods from literature regardless of the backbone model used.

Comparing both generalization protocols shows that attribute-based person retrieval primarily benefits from additional and more diverse training data.
However, there is still much room for improvement.
Considering real-world applications, a \acrshort{r1} score of 15.7\% might not be sufficient.
Even with training data from multiple domains, further research is required.
While a positive match at the first ranking position in almost every fourth case seems sufficient in some cases, the \acrshort{map} indicates that many matches occur in late ranking positions.
As a result, relevant people might be missed.
This finding indicates that current attribute classifiers can be transferred to other domains and used for retrieval but they only work well for the simple cases so far.
Besides, \acrshort{par} models seem to be better suited for generalization tasks than methods aiming at learning a cross-modal feature space using adversarial training.

\subsection{Specialization Results}
\begin{table}[t]
\centering%
\resizebox{\linewidth}{!}{ 
\begin{tabular}{@{}l|l|cc|cc|cc|cc@{}}%
\toprule%
\multirow{2}{*}{Method} & \multirow{2}{*}{Backbone} & \multicolumn{2}{c|}{PETA} & \multicolumn{2}{c|}{\acrshort{pa100k}}  & \multicolumn{2}{c|}{\acrshort{rap}} & \multicolumn{2}{c}{\acrshort{market}} \\%
\cline{3-10}%
&& mA  & F1  & mA  & F1  & mA  & F1 & mA & F1 \\%
\midrule
MsVAA\cite{sarafianos2018deep} & ResNet101 & 84.6 & 86.5 &--&--&--&--&--&--\\
VAC \cite{guo2019visual} & \acrshort{resnet50} & 83.6 & 86.2 & 79.0 & 86.8 & -- & -- & -- & -- \\  
ALM \cite{tang2019improving}  & BN-Inception & 86.3 & 86.9 & 80.7 & 86.5 & -- & -- & -- & -- \\
JLAC \cite{tan2020relation} & \acrshort{resnet50}\textsuperscript{1}  & 87.0 & 87.5 & \textcolor{blue}{\bf{82.3}} & 87.6 & -- & -- & -- & -- \\ 
VFA \cite{chen2021enhance} & \acrshort{resnet50} & 86.5 & 87.3 & 81.3 & 87.0 & -- & -- & -- & -- \\
MSCC \cite{zhong2021improving} & \acrshort{resnet50} & -- & -- & 82.1 & 86.8 & \textcolor{red}{\bf{80.2}} & 79.1 & -- & -- \\ 
SB~\cite{jia2021rethinking} & \acrshort{resnet50} & 84.0 & 86.4 & 80.2 & 87.4 & 78.5 & 78.7 & -- & -- \\
SB~\cite{jia2021rethinking}$^\dagger$ & \acrshort{resnet50} & 84.0 & 86.3 & 81.6 & 88.1 & 77.4 & 78.5 & 76.5 & 83.6 \\
SSC\textsubscript{soft} \cite{jia2021spatial} & \acrshort{resnet50} & 86.5 & 87.0 & 81.9 & 86.9 & -- & -- & -- & -- \\
\midrule
\multirow{2}{*}{Ours}  & \acrshort{resnet50} & \textcolor{blue}{\bf{87.1}} & \textcolor{blue}{\bf{87.7}} & 82.2 & \textcolor{blue}{\bf{88.5}} & 78.8 & \textcolor{blue}{\bf{80.0}} & \textcolor{blue}{\bf{79.5}} & \textcolor{blue}{\bf{85.4}} \\%
& \acrshort{cnb} & \textcolor{red}{\bf{88.4}} & \textcolor{red}{\bf{89.9}} & \textcolor{red}{\bf{84.8}} & \textcolor{red}{\bf{90.2}} & \textcolor{blue}{\bf{79.9}} & \textcolor{red}{\bf{81.0}} & \textcolor{red}{\bf{81.5}} & \textcolor{red}{\bf{87.6}} \\%
\bottomrule
\end{tabular}
} 
\caption{
\textbf{State-of-the-art \acrshort{par}} -- Our models achieve state-of-the-art results on every benchmark.
Only on \acrshort{pa100k} and \acrshort{rap} only few approaches from the literature report higher \acrshort{ma} scores.
The reason is that our models were optimized for attribute-based retrieval.
\textcolor{red}{\bf{Red}} and \textcolor{blue}{\bf{blue}} colors highlight the best and second-best results, respectively.
$\dagger$~Results were produced using the official implementation.
} 
\label{tab:sota_par}
\vspace{-0.3cm}
\end{table}
\paragraph{Ablation}
We also evaluate the regularization methods for the specialization case, i.e., training and testing data originate from the same distribution, in \cref{tab:public}.
We observed that in contrast to the generalization experiments, optimal hyperparameters and improvements by the optimizations greatly depend on the dataset.
This finding indicates that the datasets are biased and that models optimized using the same data source for training and testing may result in poor performance in real-world applications.
Also, \acrshort{re} leads to improvements in the specialization case, which hurts generalization ability, as was observed in the \acrshort{upar} experiments.
These findings clearly highlight the necessity of a general benchmark such as \acrshort{upar} consisting of data from different domains with varying image and attribute distributions.
Otherwise, developing generalizable and applicable algorithms is hardly possible.

\paragraph{Comparison with State-of-the-art}
Tables \ref{tab:sota_par} and \ref{tab:sota_retrieval} provide comparisons with the current state-of-the-art for \acrshort{par} and attribute-based person retrieval.
Regarding both tasks, our approach sets a new state-of-the-art regardless of the choice of the backbone architecture.

\begin{table}[t]%
	\centering%
	\resizebox{\linewidth}{!}{%
		\begin{tabular}{l|cc|cc|cc|cc}%
			\toprule%
			\multirow{2}{*}{Method}  & \multicolumn{2}{c|}{PETA} & \multicolumn{2}{c|}{PA100K} & \multicolumn{2}{c|}{\acrshort{rap}} & \multicolumn{2}{c}{\acrshort{market}}  \\  \cline{2-9}%
			& \acrshort{map} & \acrshort{r1} & \acrshort{map} & \acrshort{r1} & \acrshort{map} & \acrshort{r1} & \acrshort{map} & \acrshort{r1} \\ \midrule%
			DeepMAR~\cite{li2015multi} & -- & -- & -- & -- & -- & -- & 8.9 & 13.2 \\%
			DCCAE~\cite{wang2015deep} & -- & -- & 15.6 & 21.2 & - & - & 9.7 & 8.1 \\%
			2WayNet~\cite{eisenschtat2017linking} & -- & -- & 10.6 & 19.5 & - & - & 7.8 & 11.3 \\%
			CMCE~\cite{li2017identity} & -- & -- & 13.1 & 25.8 & -- & -- & 22.8 & 35.0 \\%
			AAIPR~\cite{yin2017adversarial} & -- & -- & -- & -- & -- & -- & 20.7 & 40.3 \\%
			AIHM~\cite{dong2019person} & -- & -- & -- & -- & -- & -- & 24.3 & 43.3 \\%
			SAL~\cite{cao2020symbiotic}$^\dagger$ & -- & -- & 15.0 & 22.7 & -- & -- & 29.4 & 44.4 \\%
			SB~\cite{jia2021rethinking}$^\dagger$ & 20.7 & 21.3 & 23.8 & 31.1 & 17.0 & 12.1 & 23.8 & 39.5 \\%
			ASMR~\cite{jeong2021asmr} & -- & -- & 20.6 & 31.9 & -- & -- & 31.0 & \textcolor{blue}{\bf{49.6}} \\%
			\midrule%
			Ours (\acrshort{resnet50}) & \textcolor{blue}{\bf{23.0}} & \textcolor{blue}{\bf{22.9}} & \textcolor{blue}{\bf{25.6}} & \textcolor{blue}{\bf{32.5}} &  \textcolor{blue}{\bf{19.5}} &  \textcolor{blue}{\bf{13.8}} & \textcolor{blue}{\bf{32.3}} & 45.0 \\%
		    Ours (\acrshort{cnb}) & \textcolor{red}{\bf{30.2}} & \textcolor{red}{\bf{29.7}} & \textcolor{red}{\bf{30.5}} & \textcolor{red}{\bf{39.5}} &  \textcolor{red}{\bf{21.7}} &  \textcolor{red}{\bf{15.8}} & \textcolor{red}{\bf{40.6}} & \textcolor{red}{\bf{55.4}} \\%
			\bottomrule%
	\end{tabular}
	}
	\caption{
    \textbf{State-of-the-art attribute-based retrieval} -- Our models achieve state-of-the-art results on every benchmark.
    \acrshort{par}-based and feature alignment-based methods are significantly outperformed independent of the backbone model.
    \textcolor{red}{\bf{Red}} and \textcolor{blue}{\bf{blue}} colors highlight the best and second-best results, respectively.
    $\dagger$~Results were produced using the official implementation.
    } 
	\label{tab:sota_retrieval}%
	\vspace{-0.3cm}
\end{table}

\section{Discussion of Potential Societal Impact}
The fields of \acrshort{par} and attribute-based retrieval, which encompass fashion attributes as well as attribute related to visual surveillance have several potential uses in real-world scenarios. 
Intended scenarios may include the use of retrograde \acrshort{par} and retrieval systems by law enforcement authorities to identify suspects based on recalled testimonies. 
Another application may be group detection and monitoring during large events in public transportation.
However, malicious use would imply identifying persons and groups based on their ethnicity, skin color, religion, and or cultural accessories or associated clothing. 
Mitigation strategies should exclude the integration of cultural and religious attributes into public datasets. 
Furthermore, it is currently unclear how well \acrshort{par} models differentiate between clothes and skin colors in low-resolution surveillance images, \ie, whether the model could mistake a dark-skinned person with short lower-body clothing for a person with black long lower-body clothing (\eg leggings). 
We think that our \acrshort{upar} dataset offers more diversity and lower the impact of biases in dataset. 
Further work should study latent biases in public dataset and \acrshort{upar}.

\section{Conclusions}
In this work, we proposed a unified dataset named \acrshort{upar} to allow generalization experiments for 40 attributes across four \acrshort{par} datasets along with standard evaluation schemes with 148,048 train, 30,830 validation, and 45,859 test images to measure the generalization ability of \acrshort{par} and attribute-based person retrieval approaches.
Besides, we develop a robust baseline that outperforms state-of-the-art approaches in generalization and specializiation problems of \acrshort{par} and retrieval.
Based on \acrshort{upar}, we believe the research community will develop new large-scale generalizable algorithms 
for attribute recognition and attribute-based person retrieval.

{
    \small
    \bibliographystyle{ieee_fullname}
    \bibliography{macros,main}
}



\end{document}